%% file: main.tex
\documentclass[conference]{IEEEtran}
\usepackage{times}

\usepackage[numbers]{natbib}
\usepackage{multicol}
\usepackage{amsmath}
\usepackage{amsfonts}
\usepackage{amssymb}
\usepackage[bookmarks=true]{hyperref}

\usepackage{balance}
\usepackage{multirow}
\usepackage{booktabs}
\usepackage{array}
\usepackage{boldline}
\usepackage{hhline}

\let\labelindent\relax
\usepackage{enumitem}

\usepackage{tikz}

\newcommand{\mypara}[1]{\vspace{0.1cm}\noindent\textbf{#1}}

\begin{document}

\title{Translating Natural Language to Planning Goals with Large-Language Models}

 \author{%
   Yaqi Xie$^1$, \ \  Chen Yu$^1$,  Tongyao Zhu$^1$, \ \  Jinbin Bai$^1$, \ \  Ze Gong$^1$,\ \ Harold Soh$^{1,2}$\\
   $^1$Dept. of Computer Science, School of Computing \\
   $^2$Smart Systems Institute \\
   National University of Singapore \\
   \texttt{\{yaqixie, yuchen21, tongyao.zhu, jinbin, zegong, harold\}@comp.nus.edu.sg}}

\maketitle

\input{abstract}

\IEEEpeerreviewmaketitle

\input{introduction}

\input{preliminaries}
\input{problem_statement}

\input{experiments}

\input{resultsdisc}

\input{conclusion}

\input{ack.tex}

\balance
\bibliographystyle{plainnat}
\bibliography{references}
\input{appendix}
\end{document}

%% file: abstract.tex
\begin{abstract}
Recent large language models (LLMs) have demonstrated remarkable performance on a variety of natural language processing (NLP) tasks, leading to intense excitement about their applicability across various domains. Unfortunately, recent work has also shown that LLMs are unable to perform accurate reasoning nor solve planning problems, which may limit their usefulness for robotics-related tasks. In this work, our central question is whether LLMs are able to translate goals specified in natural language to a structured planning language. If so, LLM can act as a natural interface between the planner and human users; the translated goal can be handed to domain-independent AI planners that are very effective at planning. Our empirical results on GPT 3.5 variants show that LLMs are much better suited towards translation rather than planning. We find that LLMs are able to leverage commonsense knowledge and reasoning to furnish missing details from under-specified goals (as is often the case in natural language). However, our experiments also reveal that LLMs can fail to generate goals in tasks that involve numerical or physical (e.g., spatial) reasoning, and that LLMs are sensitive to the prompts used. As such, these models are promising for translation to structured planning languages, but care should be taken in their use.
\end{abstract}

%% file: introduction.tex
\section{Introduction}

\begin{figure}
\begin{center}
\includegraphics[width=0.47\textwidth]{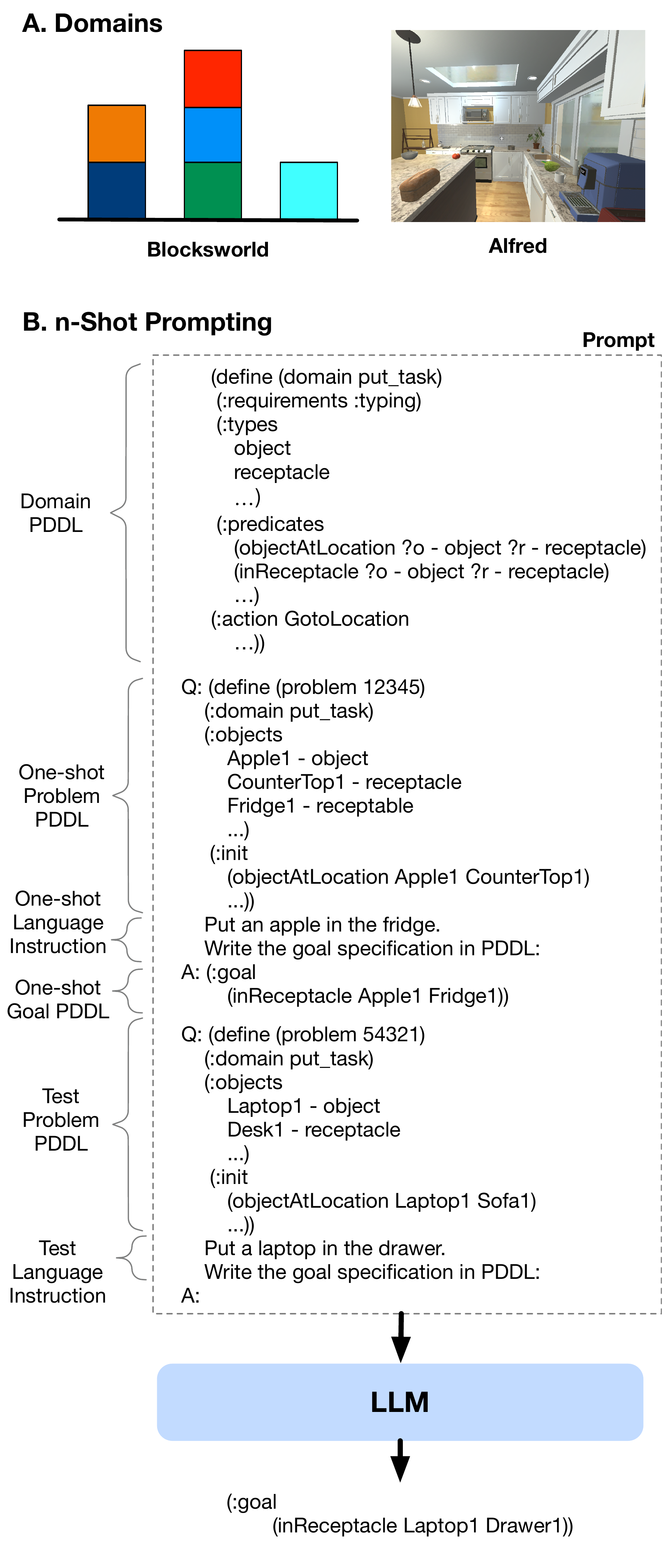}
\end{center}
\caption{This paper investigates LLM-based translation of natural language into PDDL goals. (A) Our experiments use two domains: Blocksworld and Alfred. (B) Using $n$-shot prompts ($n=1$ in this example), we find that LLMs can effectively translate English goals to PDDL when the language instruction is sufficiently specific. However, the LLM fails to consistently produce accurate goals in tasks that require formal or physical reasoning on the specified planning domain/problem.}
\label{fig:overview}
\end{figure}

The world is currently abuzz with enthusiasm over recent transformer-based models that are trained on very large amounts of text data. Despite being trained in a purely self-supervised fashion, these large language models (LLMs) have shown tremendous versatility in the types of tasks they can perform. Within robotics, recent works have examined their viability for planning and to provide high-level semantic knowledge~\cite{raman2022, pallagani2022, silver2022,ahn2022,huang2022, chen2022nlmapsaycan}. Despite their promise, recent evidence strongly suggests that LLMs are, on their own, poor planners~\cite{valmeekam2022, collins2022, mahowald2023}. This is perhaps unsurprising given that these models are trained for text completion, rather than to reason over specified domains. 

In this work, we consider the task of translation (Fig. \ref{fig:overview}). Instead of asking an LLM to plan (which is perhaps a dubious enterprise), we simply ask it to extract a planning goal from given a natural language instruction. Given the effectiveness of LLMs on natural language tasks, this translation task may appear as a natural fit and if a goal can indeed be extracted, we can leverage highly-effective classical planners to derive a plan. This modular approach  leverages the strengths of each component: the LLM provides linguistic competence, while the planner provides structured reasoning. 

However, can LLMs actually translate goals from natural language to PDDL? Translation is a non-trivial task. In the case of translation between human languages, we are keenly aware that direct literal translation is often sub-optimal. Proper translation requires one to leverage contextual information and prior knowledge. When translating to PDDL, there are additional constraints; PDDL is a formal language where small syntax errors can cause failures to parse and subsequently plan. Moreover, goal translation needs to be compatible with the given PDDL domain and problem information. 

Here, we contribute an empirical study aimed at answering the above question. Specifically, we use GPT-3.5 to translate English instructions to PDDL goals. We present results from two domains, Blocksworld~\cite{slaney2001}, which enables complex specifications, and ALFRED~\cite{shridhar2020}, which is based on real-world household environments. Our results indicate that GPT-3.5 can be an effective translator. It is able to generate  achievable goals that are consistent with the natural language instruction and the provided PDDL domain/problem specifications. In particular, we find that goals can be obtained even with ambiguous commands or those that require general commonsense knowledge. For example, when asked to generate a goal from the instruction ``\emph{Set a table for two}'', the LLM is able to furnish a complete goal specification comprising correct tableware (e.g., two cups, plates, and utensils).

Based on the above, one tempting conclusion might be that we can simply ``plug-in'' GPT-3.5 as a natural language interface to planners. However, the reality is less straightforward. Our experiments also show that LLMs can behave in counter-intuitive ways; they perform poorly on simple counting and spatial inference during translation. Example failure cases include simple commands like ``\emph{Put the pen into the box that has three books}''.
Through a subtask analysis approach, we attempt to pinpoint how these models are failing --- our analysis suggests that the LLM fails to properly understand predicate semantics in the Blocksworld domain, nor appreciates hierarchical relationships between objects in ALFRED.
We believe these results will be useful for those of us designing robot systems that leverage LLMs as they highlight potential pitfalls when using these black-box models. In summary, we show that LLMs can be remarkably potent translators due to their linguistic competence (and certainly, are more effective at translation than at planning), but at the same time, more research is needed to successfully apply them to general goal translation.

%% file: preliminaries.tex
\section{Preliminaries and Related Work}
In this section, we provide a brief overview of background material and prior work related to our goal translation. Here, we focus on enabling robots to comprehend human utterances; there is significant work in enabling robots to generate utterances (e.g., ~\cite{chen2022mirror}), which we will not delve into here. We begin with a description of Large Language Models (LLMs), followed by a short review of PDDL and machine translation to structured languages. 

\subsection{Large Language Models}

Large Language Models (LLMs) are self-supervised nonparametric models that are trained on a very large corpus of text data. They typically have a considerable number of parameters, ranging from hundreds of millions \cite{devlin2018bert} to billions \cite{brown2020language}. Examples of LLMs include Google's PaLM~\cite{chowdhery2022palm} and LaMDA~\cite{thoppilan2022lamda}, and OpenAI's GPT models~\cite{brown2020language}. LLMs are typically trained to perform language completion, e.g., to predict the likelihood of the next tokens conditioned upon some text. Nevertheless, by providing suitable prompts, LLMs can be made to perform surprisingly well on many NLP tasks~\cite{brown2020language,devlin2018bert,raffel2020exploring}. In particular, when given a few task examples (called a few-shot prompt), LLMs can quickly be quickly adapted for specific use-cases~\cite{huang2022towards,li2022pre,huang2022language,ahn2022}. Our experiments in this work are on GPT-3.5 variants with zero-shot prompts (a query but no examples) and $n$-shot prompts (where $n$ examples are provided along with the query). 

\subsection{Planning Domain Definition Language (PDDL)}

The Planning Domain Definition Language (PDDL)~\cite{fox2003pddl2} is a family of STRIPS-style languages that define a planning problem. In the following, we give a brief overview of PDDL and refer interested readers to comprehensive guides~\cite{geffner2013concise}. A PDDL planning problem is split into two parts~\cite{pddlwiki}:
\begin{itemize}
    \item A domain that defines the ``universal'' aspects of a problem, i.e., the elements that are present in all specific problem instances such as object types, predicates, and available actions. An action specifies a change to the state of the world and is typically structured into three parts: its parameters, preconditions, and effects. %
    \item A problem which describes the exact instance or situation that we aim to solve. It specifies the objects that exist in the world and an initial state. A key element of interest in this work is the \emph{goal}. The goal is a set of logical expressions of predicates that should be satisfied. 
\end{itemize}
State-of-the-art AI planners such as Fast Downward~\cite{helmert2006fast} take as input a PDDL domain and problem, and output a plan that results in goal satisfaction. 

\subsection{Machine Translation from Natural Language to Structured Languages}

Goal translation is a specific form of machine translation to structured languages. Direct linguistic parsing and mechanical substitution often fail to produce useful goal translations --- natural language goals can be stated in various ways and may be incomplete, requiring common sense reasoning and world knowledge to furnish missing details. For example, translating ``\emph{make a tomato sandwich}'' into a PDDL goal requires a robot to know the composition of a tomato sandwich (that the tomato is sliced and between two slices of bread). Moreover, the translator has to be linguistically competent; it has to generate a goal that is syntactically correct and consistent with the domain specification.  

There is significant prior work in structured programming code generation using natural language and recent work applying LLMs towards programming languages~\cite{chen2021evaluating}. Within the context of natural language understanding for robot planning and decision-making, early works applied classical NLP tools and predefined rules, e.g., for extracting objects and actions to generate a PDDL domain~\cite{steinert2020planning}. Later works used non-parametric flexible models such as deep neural networks for various tasks involving the extraction of PDDL models/plans from natural language~\cite{steinert2020planning,miglani2020nltopddl,feng2018extracting,simonnatural} and to translate natural language commands to LTL goal specifications \cite{gopalan2018sequence,patel2020grounding}. Recent work has applied LLMs for translation to linear temporal logic (LTL) formulae~\cite{fuggitti2023nl2ltl,liu2022lang2ltl}. In this work, we focus on translation to PDDL. Very recent work has shown that LLMs are unable to reliably generate plans in PDDL domains~\cite{olmo2021gpt3,silver2022}. 

Our work is related to very recent work by Collins et al.~\cite{collins2022}, which showed that a ``parse-and-solve'' model --- using the LLM to translate a natural language goal into PDDL that can be processed by a planner --- was more effective than direct LLM planning. This parse-and-solve approach is intuitive and appealing in that it separates goal understanding from planning. PDDL goals are interpretable and it is possible to perform constraint checking and verification of the goal before passing it to a planner. While initial results are promising~\cite{collins2022}, it remains unclear how effective LLMs are at translation. In our study, we examine this question in a comprehensive manner using various tasks (e.g., partially-specified goals with ambiguity) in two domains. %

%% file: problem_statement.tex
\section{Problem Statement: Goal Translation with Large-Language Models}
 \label{sec:problemstatement}
 
In this work, we investigate the potential of state-of-the-art large-language models (LLMs) --- specifically, variants of GPT-3.5 --- for \emph{n-shot} translation.
Given a query prompt $q$, we are interested in obtaining a goal state $g$ specified in planning language (PDDL). We assume the goal $g$ can be sampled from a posterior distribution,
\begin{align}
g \sim p(g|q)
\label{eq:goalposterior}
\end{align} 
where the query $q = (q_m, q_e, q_n)$ comprises a description of the domain $q_m$ (written in PDDL) and the desired goal state in natural language $q_n$. $q$ also contains $q_e$ which comprises $n$ examples of correct translations; this form of prompting is highly effective for inducing desired outputs from LLMs and has been used in prior work on LLM planning and translation (e.g., \cite{silver2022}). An example prompt used in our work is shown in Fig. \ref{fig:overview}.   

It is generally challenging to precisely pinpoint the exact reason for observed performance since state-of-the-art LLMs are large transformer-based models. Moreover, these models remain closed-sourced with limited API access.
Prior works have tended to focus on success rates that illustrate under what circumstances (or what tasks) these LLMs can plan or translate~\cite{collins2022}. We will follow suit and also report success rates as a primary metric. However, success rates only give us a glimpse into the capabilities and limitations of LLMs. It remains an open question as to \emph{what} underlies a model's specific performance on a task (or lack thereof). 

\subsection{Task Types}
To better understand under what circumstances LLMs can effectively translate, we design tasks that require different underlying capabilities. These tasks range from simple fully specified tasks (where the goal configuration is specified unambiguously) to partially-specified instructions where inference or commonsense reasoning is required. We delay the precise description of our task types to Sec. \ref{sec:experimentsetup} on the experimental setup.

\subsection{Translation Subtasks}
We will attempt to understand the limitations of the LLM via an analytical approach that breaks apart the translation task and the conditional probability distribution in Eqn. (\ref{eq:goalposterior}). In particular, our translation task can be cognitively decomposed into three key steps:
\begin{enumerate}
    \item \textbf{Domain Understanding:} the LLM has to parse and comprehend the PDDL domain specification;
    \item \textbf{Goal Inference:} the LLM has to infer the intended goal given the natural language utterance and domain knowledge;
    \item \textbf{PDDL Goal Specification:} the LLM has to output the inferred goal in the correct PDDL syntax.
\end{enumerate}
In a similar vein, we can factorize Eqn. \ref{eq:goalposterior} into the following:
\begin{align}
p(g|q) = \iint p(g|z,m,q)p(z|m,q)p(m|q)dm dz
\label{eq:factorized}
\end{align}
where $m$ represents domain information that can be inferred from the query $q$, i.e., the objects, actions, and predicates within the PDDL domain specification. The distribution $p(z|m,q)$ represents the posterior over a ``latent'' goal $z$ obtained via inference over the domain $m$ and natural language goal within the query. 
This latent goal $z$ needs to be transcripted into PDDL goal $g$. While the LLM may not explicitly represent the distributions in (\ref{eq:factorized}) above, the factorization applies to any conditional distribution (\ref{eq:goalposterior}). This factorization, along with the decomposition of the translation task, above provides a starting point for our analytical approach.  
Potentially, we can investigate possible causes by changing parts of Eqn. (\ref{eq:factorized}) via careful modifications to the conditioning prompt $q$. Specifically, to investigate:
\begin{enumerate}
    \item \textbf{Domain Understanding:} we focus on $p(m|q)$ by keeping the PDDL domain specification $q_m$ the same, but changing $q_n$ to directly ask about elements in the domain (e.g., the objects and predicates);
    \item \textbf{Goal Inference:} potentially, the model has correctly inferred $p(z|m,q)$ but fails to generate the goal in PDDL. To better distinguish these situations, we can ask the model to generate $z$ in different languages (e.g., natural language or Python). If the model is able to output a valid goal in a different syntax, it suggests that the model has correctly inferred $z$;
    \item \textbf{PDDL Goal Specification:} likewise, we can examine if the model is unable to correctly output the goal specification in PDDL $p(g|z,m,q)$ if the goal inference is correct (the model is able to generate $z$ in a different language) but the PDDL output is wrong.
\end{enumerate}
Note that care should be taken in interpreting the outcomes of such prompt modifications; we may inadvertently change different factors simultaneously. In addition, we are always reliant on the syntax output distribution and thus, cannot completely isolate the individual subtasks. Nevertheless, we assume that by restricting our changes to only certain parts of the prompt, we are limiting the changes to the distributions. 
Then, by correlating the results of the above tests to the nature of the translation task, we can better understand the failure points in the model. We will further discuss the application of this analysis in Section \ref{sec:ResultsAndDiscussion_error}.

%% file: experiments.tex
\section{Experimental Setup}
\label{sec:experimentsetup}

This section describes our experimental setup, including the domains, tasks, and evaluation methodology. 

\subsection{Domains}

\mypara{Blocksworld}
is a classical planning domain, widely used for education and research purposes due to its simplicity in specification and complexity in optimal planning. Blocksworld problems describe vertical spatial relations between objects on a table. A goal state specifies objects in a certain order. %
In our setup, the objects are all colored blocks, similar to Valmeekam et al.~\cite{valmeekam2022}. For most tasks, every block has a unique color, e.g., \texttt{red\_block}. However, there are certain tasks where multiple blocks have the same color and are distinguished by an additional index suffix in the problem definition, e.g., \texttt{red\_block\_1}.

\mypara{ALFRED-L} is based on Action Learning From Realistic Environments and Directives (ALFRED)~\cite{shridhar2020}, a household domain for learning to map instructions in natural language and egocentric vision to action sequences. We chose ALFRED as it is consistent with our goal of translating from natural language and is representative of possible real-world applications. %
In our setup, we use three basic classes of objects in ALFRED, namely, \texttt{object}, \texttt{receptacle}, and \texttt{agent}. An \texttt{object} is a base class for a movable item. A \texttt{receptacle} may contain objects, and an \texttt{agent} is able to execute actions on objects. A \texttt{receptacle\_object} is a subclass of \texttt{receptacle} and \texttt{object}, and denotes a movable object that can contain another object. We reduced the original domain and problem specification so that it would fit within the LLM's query window length by removing unused predicates and combining \texttt{inReceptacle} and \texttt{inReceptacleObject}. With these classes, we construct different scenes with randomly initialised objects.
For each scene, we also initialise one agent to allow the goal state to contain predicates involving an agent. Our scenes include: 
\begin{enumerate}
    \item \textbf{Kitchen:} containing foods (e.g., \texttt{Apple}, \texttt{Potato}), utensils (e.g., \texttt{Knife}, \texttt{Fork}), and receptacles (e.g., \texttt{SinkBasin} and \texttt{Fridge}).
    \item \textbf{Living room:} containing items (e.g., \texttt{FloorLamp}, \texttt{Laptop} and \texttt{RemoteControl}) and receptacles (e.g., \texttt{Sofa} and \texttt{Drawer}). 
    \item \textbf{Bedroom:} containing items like \texttt{AlarmClock}, \texttt{Book}, and \texttt{CD}, and receptacles like \texttt{Drawer} and \texttt{Bed}. 
\end{enumerate}

\subsection{Task Design}

A typical configuration in a Blocksworld problem consists of one or more \textit{stacks}, where each stack is a sequence of blocks placed one on top of the other. In ALFRED-L, the tasks comprise interacting with objects (e.g., moving, slicing) to arrive at a goal configuration (e.g., a sandwich). By describing the desired configuration with different levels of specificity or  constraints, we can derive problems tasks that require one or more of the following skills: 
\begin{enumerate}
    \item \textbf{Linguistic competence:} Able to parse the text correctly. The model has to recognize entities that are referred to in the text and understand their relations that are encoded with verbs and prepositions (and predicates in PDDL).
    \item \textbf{Object association:} Able to associate objects in PDDL problems with entities in the natural language text. In cases where blocks may have the same color, the model needs to ground an ambiguous reference like ``a red block'' to a concrete PDDL symbol like \texttt{red\_block\_1}. Likewise, in ALFRED-L, the model would like to ground an item like ``book'' to \texttt{Book1}.
    \item \textbf{Numerical reasoning:} Able to count objects and perform simple arithmetic reasoning. For example, the model needs to build a stack with a certain amount of blocks, or distribute blocks evenly in every stack. 
    \item \textbf{Physical reasoning:} Able to understand physical law and relationships between objects. For example, the model needs to infer the spatial relations from a sequence representation of blocks in a stack. In ALFRED-L, the model has to infer whether one object contains another.
    \item \textbf{World Knowledge:} Able to infer valid configurations from instructions that involve semantically rich object properties, especially those that are not explicitly specified in the domain. In ALFRED-L, world knowledge includes commonsense relationships between objects (e.g., an apple is a fruit; ice cream is usually kept in the fridge)%
\end{enumerate}

\input{tbl_task.tex}

The tasks we crafted are shown in Table \ref{tbl:task}, along with the skills that the tasks were designed to test. The number of instances for each task varied but in general, for Blocksworld tasks, we tested problems with 4,8 and 12 blocks (100 instances with differing configurations each). The ExplicitStacks-II task has a large number of tasks as we also varied the ordering specification (``\emph{bottom to top}'' and ``\emph{top to bottom}''). The ALFRED-L tasks comprised 100 instances for each task (with different goal objects and initial states), except the tasks with the broadest instructions (i.e., CutFruits, PrepareMeal, IceCream, SetTable2, CleanKitchen). For almost all the tasks, we used one-shot prompting ($n=1$) to limit the prompt length. We also tested zero-shot prompts but the performance was poorer in those cases (see Supplementary Material).

\subsection{LLM Models}

We used variants of GPT-3.5 for our experiments, specifically \texttt{code-davinci-002} for all Blocksworld tasks and \texttt{text-davinci-003} for ALFRED-L. These two models were chosen because they achieved the best performance in initial translation experiments. 

\subsection{Evaluation Methods}

\mypara{Translation Success Rate.} As stated in Sec. \ref{sec:problemstatement}, we use  success rates as a metric for evaluating the translation performance of the LLM. For both domains, we had \emph{strict} and \emph{loose} criteria for success, similar to prior work~\cite{silver2022}. The strict criteria required that the LLM produce a goal that was acceptable to a planner and did not needlessly change predicates. In Blocksworld, a goal stack was correct under the \emph{strict} criteria if (i) the block order was correct, (ii) the bottom-most block is stated as being on table and (iii) the top-most block is stated as clear (having nothing on top of it). The \emph{loose} metric captures instances where the planner did not satisfy either (ii) or (iii), i.e., the ordering was correct but it failed to consistently apply the \texttt{clear} and \texttt{ontable} predicates. For ALFRED-L tasks, we manually designed rules to check whether the generated goal satisfies each task instruction, e.g., for SetTable2, that two sets of tableware were placed on the table. The loose criteria in this case ensure that corresponding predicates must be true and the strict criteria additionally check that no unnecessary predicates are specified (e.g., for SetTable2, that no items are in the sink). 

For Blocksworld, we designed additional checks that indicate what was wrong with the goal stack:
\begin{itemize}
    \item \textbf{A Domain/Syntax error check} evaluates if the goal complies with the PDDL context. The parser reports an error if the goal has a wrong syntax or includes an object that doesn't exist in the world. 
    \item \textbf{A Physical error check} tests if the predicates form a stack that is physically possible to construct (if the domain/syntax check is passed). Typical errors include having more than one block on top of the same block, or that a block sits above and below another block at the same time.
\end{itemize}

\mypara{Subtask Scores.} In addition to the metrics above, we conducted tests on the LLMs ability to perform domain understanding and goal inference. In Blocksworld, our domain understanding test comprised five basic queries:
\begin{itemize}
    \item \textbf{Object extraction:} ``\emph{List all the objects that appear in the PDDL problem definition}''.						
    \item \textbf{Color-based Object extraction:} the query is similar to above except we add a constraint ``\emph{that has the color [color]}''.
    \item \textbf{On Predicate:} ``\emph{Determine whether the [object A] is on the [object B] in the initial state. Answer with yes or no}''.						
    \item \textbf{Clear Predicate:} ``\emph{Determine whether the [object] is on the table in the initial state. Answer with yes or no.}''  		
    \item \textbf{OnTable Predicate:} ``\emph{Determine whether there is nothing on the top of the [object] in the initial state. Answer with yes or no}''.						
\end{itemize}
The score is an average of how many of the above questions the LLM answered correctly. To test goal inference ability, we tasked the LLM to generate the stacks as Python lists. This was inspired by the observation in preliminary trials that on some tasks, the LLM would successfully output Python lists even though the PDDL goal was incorrect. We acknowledge this is an imperfect test but is one of the few available to us given the constraints of working with black-box LLMs. In ALFRED-L, the domain and goal inference questions were tailored for each task (see Supplementary Material). For example, for MoveToCount2, the relevant queries were:
\begin{itemize}
    \item \textbf{Domain understanding:} ``\emph{Which box has two [target types] in the initial state?}''
    \item \textbf{Goal Inference:} ``\emph{Move [moved object] to the box with two [target types]. Which object should we move the [moved object] into?}''
\end{itemize}

%% file: tbl_task.tex
\begin{table*}[]
\centering
\label{tbl:task}
\caption{Task Descriptions. Legend: LC (Linguistic Competence), OA (Object Association), NR (Numerical Reasoning), PR (Physical reasoning, including spatial), WK (World Knowledge)}
\begin{tabular}{llp{0.16\linewidth}clllllp{0.23\linewidth}}
\hline\hline
                &                    &                                                                                              & \multicolumn{1}{l}{}                      & \multicolumn{5}{c}{\textbf{Primary Skill(s) Tested}}                                                                                                                    &                                                                                                                                                                                                                                                 \\
\textbf{Domain} & \textbf{Task Name} & \textbf{Task Description}                                                                    & \multicolumn{1}{l}{\textbf{\# Instances}} & \multicolumn{1}{c}{\textbf{LC}} & \multicolumn{1}{c}{\textbf{OA}} & \multicolumn{1}{c}{\textbf{NR}} & \multicolumn{1}{c}{\textbf{PR}} & \multicolumn{1}{c}{\textbf{WK}} & \textbf{Example}                                                                                                                                                                                                                                \\ \hline
BlocksWorld     & ExplicitStacks     & Stack configurations specified with no object ambiguity                                      & 300                                       & \checkmark                            &                            &                            &                            &                            & \textit{Build two stacks. In the first stack, the black block is on the table, the green block is on top of the black block, the violet block is on top of the green block, and there is nothing on the violet block. In the second stack, ...} \\
                & ExplicitStacks-II  & Stack configurations specified with no object ambiguity (alternative ordering specification) & 600                                       & \checkmark                            &                            &                            & \checkmark                            &                            & \textit{Build four stacks. In the first stack, there are the gold block, and the red block from bottom to top. In the second stack, ...}                                                                                                        \\
                & BlockAmbiguity     & Stack configurations specified with object ambiguity (multiple blocks with the same color)   & 200                                       & \checkmark                            & \checkmark                            &                            &                            &                            & \textit{Create two stacks. In the first stack, a red block is on top of a red block, and the yellow block is on top of a red block....}                                                                                                         \\
                & NBlocks            & One stack with N blocks                                                                      & 300                                       & \checkmark                            &                            & \checkmark                            & \checkmark                            &                            & \textit{Create a stack that contains two blocks.}                                                                                                                                                                                               \\
                & KStacks            & K stacks with the same number of blocks                                                      & 300                                       & \checkmark                            &                            & \checkmark                            & \checkmark                            &                            & \textit{Using all blocks specified in the problem, create exactly two stacks that are of the same height.}                                                                                                                                                                                 \\
                & PrimeStack         & One stack with a prime number of blocks                                                      & 200                                       & \checkmark                            &                            & \checkmark                            & \checkmark                            &  \checkmark                          & \textit{Make a stack with a prime number of blocks.}                                                                                                                                                                                            \\
                & KStacksColor       & K stacks where each stack comprises blocks with the same color                               & 200                                       & \checkmark                            & \checkmark                            & \checkmark                            & \checkmark                            &                            & \textit{Using all blocks specified in the problem, build K stacks where each stack comprises blocks with the same color.}                                                                                                                                                                                          \\ \hline
ALFRED-L          & ExplicitInstruct   & Fully-specified instruction                                                                  & 100                                       & \checkmark                            &                            &                            &                            &                            & \textit{Put a sliced apple on the plate.}                                                                                                                                                                                                       \\
                & MoveSynonym        & Move an object to a receptacle replaced with its synonym.                                    & 100                                       & \checkmark                            & \checkmark                            &                            &                            &                            & \textit{Put Pencil1 on a couch.} (Context: Name Sofa1 is used in the objects declaration.)                                                                                                                                                      \\
                & MoveNextTo         & Move an object next to another object.                                                       & 100                                       & \checkmark                            &  \checkmark                          &                            & \checkmark                            &                            & \textit{Put Book5 next to Book4. Do not move Book4.}                                                                                                                                                                                            \\
                & MoveToCount2       & Move an object to a receptacle containing 2 other objects.                                   & 100                                       & \checkmark                            & \checkmark                            & \checkmark                            & \checkmark                            &                            & \textit{Move KeyChain1 to the box with two books}                                                                                                                                                                                               \\
                & MoveToCount3       & Move an object to a receptacle containing 3 other objects.                                   & 100                                       & \checkmark                            & \checkmark                            & \checkmark                            & \checkmark                            &                            & \textit{Move KeyChain1 to the box with three books}                                                                                                                                                                                             \\
                & MoveToMore         & Move an object to the receptacle with more objects.                                          & 100                                       & \checkmark                            & \checkmark                            & \checkmark                            & \checkmark                            &                            & \textit{Move KeyChain1 to the box with more books}                                                                                                                                                                                              \\
                & MoveNested         & Move an object to a "nested" object                                                          & 100                                       & \checkmark                            & \checkmark                            &                            & \checkmark                            & \checkmark                            & \textit{Put KeyChain1 on the sofa with Book1. Do not put it in box.}  (Example Context: Book1 is in Box1, which is on Sofa1.)                                                                                                                           \\
                & MoveNested2        & Move object to a two-layer nested object                                                     & 100                                       & \checkmark                            & \checkmark                            &                            & \checkmark                            & \checkmark                            & \textit{Put KeyChain1 on the sofa with Book2. Do not put it in box.}  (Example Context: Book2 is in Box3, which is in Box4. Box4 is on Sofa2.)                                                                                                          \\
                & CutFruits          & Cut some fruits and put them on the plate.                                                   & 20                                        & \checkmark                            & \checkmark                            &                            &                            & \checkmark                            & \textit{Cut some fruits and put them on the plate.}                                                                                                                                                                                             \\
                & PrepareMeal        & Prepare a meal.                                                                              & 20                                        & \checkmark                            & \checkmark                            & \checkmark                            & \checkmark                            & \checkmark                            & \textit{Prepare a meal.}                                                                                                                                                                                                                        \\
                & IceCream           & Put the ice-cream where it belongs.                                                          & 20                                        & \checkmark                            &  \checkmark                            &                            &                            & \checkmark                            & \textit{Put the ice-cream where it belongs.}                                                                                                                                                                                                    \\
                & SetTable2          & Set the table for two persons.                                                               & 20                                        & \checkmark                            & \checkmark                            & \checkmark                            & \checkmark                            & \checkmark                            & \textit{Set the table for two persons.}                                                                                                                                                                                                         \\
                & CleanKitchen       & Clean up the kitchen.                                                                        & 20                                        & \checkmark                            & \checkmark                            &                            & \checkmark                            & \checkmark                            & \textit{Clean up the kitchen.}                                                                                                                                                                                                                  \\ \hline\hline
\end{tabular}
\end{table*}

%% file: resultsdisc.tex
\section{Results and Discussion}
\label{sec:ResultsAndDiscussion}

In the following, we summarize our main findings, starting with an analysis of the translation success rates on the different tasks, followed by a discussion of the LLM's performance on the subtasks described in Sec. \ref{sec:problemstatement}. In general, the LLM had excellent performance on tasks that has unambiguously specified goals, but the performance was mixed on tasks with goals that were only partially-specified~\footnote{We also tested a baseline translator similar to Steinert et al.~\cite{steinert2020planning} which uses traditional NLP techniques, but this method was far poorer than the LLM in initial experiments.}. Nevertheless, performance was well above chance and in all of the tasks, preliminary tests showed that LLM failed to plan (planning success rates were 0\% for most tasks, similar to findings in prior work~\cite{valmeekam2022}). 

\subsection{Success Rate Analysis} 

The success rates in Table \ref{tbl:results} show that the \textbf{LLM can directly translate explicitly and unambiguously specified goals in natural language to PDDL}, with some caveats discussed below. In ExplicitStacks and ExplicitInstruct --- where the natural language instructions precisely state the desired configuration ---- goal correctness was close to 100\%. For ExplicitStacks, goal translation performance was sustained even when the number of blocks in the domain and desired number of stacks were varied, suggesting that the LLM is able to generalize across initial domain configurations. Likewise, for ExplicitInstruct in the ALFRED-L domain, the LLM achieved perfect goal translation across multiple objects in the different scenes.

\input{tbl_results.tex}

\begin{table}
  \centering
  \caption{Success Rates Depending on the Size of the Prompt Example}
  \begin{tabular}{c|ccc}
    \hline 
    \hline 
     & \multicolumn{3}{c}{\textbf{One-Shot Example}}\\
     \textbf{Target Goal}   &   \textbf{4 blocks}   &   \textbf{8 blocks} &  \textbf{12 blocks} \\
    \midrule
    \textbf{4 blocks}  & 67\%                  & 75\%                   & 100\%     \\
    \textbf{8 blocks}  & 52\%                  & 77\%                   & 100\%      \\
    \textbf{12 blocks} & 32\%                  & 80\%                   & 99\%  \\
    \hline 
    \hline 
  \end{tabular}
  \label{tbl:one-shot}
\end{table}

\input{tbl_error_new.tex}

However, our results also show that \textbf{LLM is sensitive to the natural language goal prompt}. For example, ExplicitStacks-II also unambiguously specifies the stack configurations, e.g., ``\emph{... In the first stack, there are the magenta block, the red block, and the yellow block, from bottom to top. In the second stack, ...}'' with the difference being that the blocks are stated from ``\emph{bottom to top}'' or ``\emph{top to bottom}''. However, performance for this task was only $\approx 50\%$. Further analysis showed that the model ordered the blocks incorrectly when the natural language instruction was stated from ``top to bottom'', indicating an ordering bias in the PDDL generation. A different, but related, phenomenon was observed on the MoveNextTo task in the ALFRED-L domain. Here, the LLMs were largely capable of repositioning one object adjacent to another but if the constraint ``\emph{do not move [Object B]}'' was dropped, the LLMs specified a goal that \emph{relocates both} objects to the one location. Although this is technically correct, it is unexpected and may violate the intentions of a user. The above results indicate that the LLM was able to linguistically reason about the goal statement (together with the specified PDDL domain), but also illustrate biases in the translation process.

The LLM's performance was mixed on the tasks that involve \emph{partial specification} of the goal. As previously mentioned, these tasks require either inference or commonsense knowledge to solve. In general, we found that \textbf{the LLM was able to apply commonsense reasoning to ``fill the gaps'' in the natural language goal specification and demonstrated a degree of elementary numerical/physical reasoning}. This is most apparent in the ALFRED-L tasks in Tbl. \ref{tbl:results}, the LLM was able to identify object types (e.g., fruits), and synonyms (e.g., couch and sofa, timepiece and watch), apply elementary spatial reasoning (next to) and was aware of common relationships (e.g., that ice cream is generally kept in a fridge). It was also able to translate the ambiguous goals statements in the SetTable2 and CleanKitchen tasks to a good degree. 

However, \textbf{the LLM displayed poorer competence on tasks that involve nontrivial numerical reasoning and physical world inference}. Our results support prior findings~\cite{valmeekam2022,collins2022,mahowald2023} that LLMs have difficulty with reasoning-related tasks. We found the ALFRED-L tasks that involve counting objects (MoveCount2, MoveCount3) and spatial relationship reasoning (MoveNested, MoveNested2) were more challenging for the LLM. Errors increased as the task became more difficult (from two to three objects in the receptacle and from one-level to two-level nesting). Similarly, in Blocksworld, performance for NBlocks and KStacks, which involved numerical reasoning, was $\approx 50-60\%$. Here, the LLM tended to blindly copy the prompt example (e.g., if the prompt example included a goal with 2 stacks, the LLM would output a PDDL goal for 2 stacks even though the natural language goal specified 4 stacks). In general, we found \textbf{the LLM required a sufficiently good example to generalize well}. Table \ref{tbl:one-shot} shows that the number of blocks used on the one-shot example significantly affects the performance on target tasks and larger examples generally led to better performance. 

Somewhat surprisingly, the lowest scores in Blocksworld are associated with BlockAmbiguity and KStacksColor; these two problems require the LLM to associate objects based on their color and we had apriori expected the LLM to be capable of such associations and perform well on this task. In BlockAmbiguity, in most cases, the blocks specified in the goal were correct but with an incorrect ordering, with errors occurring near the \texttt{ontable} predicate. For KStacksColor, the LLM tended to generate goal stacks that were physically valid, but either (i) one of the stacks would not be of a single color, or (ii) built the wrong number of stacks. Likewise, performance on the MoveSynonym task was subpar.

\subsection{Subtask Analysis}
\label{sec:ResultsAndDiscussion_error}

Our subtask analysis results are summarized in Tbl. \ref{tbl:erroranalysis}. Here, we conducted separate tests for domain understanding and goal inference, and compare the scores for each subtask when the overall translation task was successful versus when it failed. By examining the change in the performance on each subtask, we aim to piece together if errors were mainly caused by the (i) model failing to properly parse/understand the domain, (ii) errors in goal inference, and/or (iii) inability to properly generate PDDL. Performance on either/both domain understanding and goal inference subtasks generally fell when translation failed, which supports our structured analysis and indicates that our tests are reasonable indications of the LLM's underlying capabilities.

We noted that the Blocksworld domain understanding scores were relatively low compared to ALFRED-L, even when the translation was successful. 
Further analysis showed that the LLM was able to perform object extraction from the PDDL (with lower performance when extracting based on color), but failed to answer predicate-related queries; the success rate on these queries was $\approx$ 40-60\%. \textbf{This suggests that the LLM did \emph{not} understand the physical world semantics of the predicates}. Instead, it leveraged on its linguistic abilities to construct goal stacks using the prompt example. This potentially explains (i) why the example prompt was key to performance in the Blocksworld domain and (ii) the nature of the errors in the BlockAmbiguity and KStacksColor domains\footnote{Interestingly, for these two problems, we noted that Python list generation was substantially poorer than PDDL generation. For KStacksColor, this was partially due to overfitting on the example prompt.}.

For ALFRED-L, for certain tasks (MoveToCount2, MoveToMore, MoveToNested), the model still retained similar domain understanding scores in the failure condition, but achieved poorer goal inference scores. This suggests \textbf{the LLM was able to parse the PDDL file to obtain relevant information, but was unable to integrate the information to arrive at a suitable goal}. For example, on the MoveToMore task, the LLM was able to infer which box had more items (e.g., books), but failed to make the seemingly simple inference to place the target item in that box. On MoveToCount3 and MoveToNested2, the model's domain understanding score fell; it was no longer able to consistently identify which item contained 3 objects or was located within another object. As expected, goal inference scores also fell. Our analysis indicates that the LLM's ability to fundamentally reason about the domain is limited (in this case,  numerical and hierarchical reasoning), leading to errors in translation.

\subsection{Discussion: Other Findings}

The above results show that LLMs have significant linguistic capabilities and as such, are sometimes able to mask their limitations (e.g., when translating in the BlocksWorld domain). We see that the overall performance of LLM is impressive on common tasks on ALFRED, presumably due to the associations learnt from the huge amount of text data it has consumed. 

We also noted that the LLM tended to presume that the given instructions were valid. For example, when asked to ``\emph{Put two oranges on the table}'' and no oranges present in the PDDL specification, the LLMs would simply create two orange instances (e.g., \texttt{Orange\_1} and \texttt{Orange\_2}) along with the goal specification. When asked to ``\emph{Put a sliced plate on the table}'', the model added a predicate that the plate should be sliced, again suggesting linguistic competence but a lack of reasoning. Interestingly, when asked directly whether a plate can be sliced, the LLMs stated ``No'', which coincides with our common-sense knowledge. The LLM appears to be able to infer the validity of some instructions, but they do not perform this validation actively. When we added ``\emph{ask a question if it is not achievable.}'', the LLM would respond correctly that oranges were not present but not that the plate was not sliceable. In short, responses were inconsistent and varied depending on the objects in question.

%% file: tbl_results.tex
\begin{table}[]
\centering
\caption{Success Rates on the Goal Translation Tasks}
\label{tbl:results}
\begin{tabular}{llrr}
\hline\hline
                &                    & \multicolumn{2}{c}{\textbf{Success Rate (\%)}}                           \\
\textbf{Domain} & \textbf{Task Name} & \multicolumn{1}{c}{\textbf{Loose}} & \multicolumn{1}{c}{\textbf{Strict}} \\ \hline
BlocksWorld     & ExplicitStacks     & 99.67                             & 98.67                              \\
                & ExplicitStacks-II  & 58.00                              & 52.17                               \\
                & BlockAmbiguity     & 18.00                              & 14.50                               \\
                & NBlocks            & 62.00                              & 57.33                               \\
                & KStacks            & 55.00                              & 55.00                               \\
                & PrimeStack         & 87.00                              & 87.00                               \\
                & KStacksColor       & 19.00                              & 19.00                               \\ \hline
Alfred          & ExplicitInstruct   & 100.00                             & 100.00                              \\
                               & MoveSynonym        & 100.00                              & 100.00                               \\
                & MoveNextTo         & 99.00                              & 99.00                               \\
                & MoveToCount2       & 96.00                              & 96.00                               \\
                & MoveToCount3       & 74.00                              &74.00                               \\
                & MoveToMore         &86.00                              & 86.00                               \\
                & MoveNested         & 59.00                              & 59.00                               \\
                & MoveNested2        & 53.00                              & 53.00 \\
                & CutFruits          & 100.00                             & 100.00                              \\
                & PrepareMeal        & 90.00                              & 60.00                               \\
                & IceCream           & 80.00                              & 75.00                               \\
                & SetTable2          & 100.00                             & 100.00                              \\
                & CleanKitchen       & 90.00                              & 35.00                               \\ \hline\hline
\end{tabular}
\end{table}

%% file: tbl_error_new.tex
\begin{table*}[htbp]
  \centering
  \caption{Performance on Subtasks (Higher Scores indicate Better Performance)}
    \begin{tabular}{rrrrrr}
    \toprule
    \toprule
          &       & \multicolumn{2}{c}{\textbf{Baseline Score}} & \multicolumn{2}{c}{\textbf{Change in Score}} \\
          &       & \multicolumn{2}{c}{\textbf{ (Successes Only)}} & \multicolumn{2}{c}{\textbf{ (Failures Only)}} \\
    \multicolumn{1}{p{5.915em}}{\textbf{Domain}} & \multicolumn{1}{p{7.335em}}{\textbf{Task Name}} & \multicolumn{1}{c}{\textbf{Domain Und.}} & \multicolumn{1}{c}{\textbf{Goal Inference}} & \multicolumn{1}{c}{\textbf{Domain Und.}} & \multicolumn{1}{c}{\textbf{Goal Inference}} \\
    \midrule
    \midrule
    \multicolumn{1}{p{5.915em}}{BlocksWorld} & \multicolumn{1}{p{7.335em}}{ExplicitStacks} & 80.00 & 100.00 & \textcolor[rgb]{ 1,  .6,  0}{0.00} & \textcolor[rgb]{ 1,  .6,  0}{0.00} \\
          & \multicolumn{1}{p{7.335em}}{ExplicitStacks-II} & 73.33 & 95.53 & \textcolor[rgb]{ .22,  .463,  .114}{1.48} & \textcolor[rgb]{ .8,  0,  0}{-13.30} \\
          & \multicolumn{1}{p{7.335em}}{BlockAmbiguity} & 68.89 & 29.63 & \textcolor[rgb]{ .8,  0,  0}{-5.11} & \textcolor[rgb]{ .8,  0,  0}{-28.47} \\
          & \multicolumn{1}{p{7.335em}}{NBlocks} & 67.69 & 96.51 & \textcolor[rgb]{ .8,  0,  0}{-4.07} & \textcolor[rgb]{ .8,  0,  0}{-2.76} \\
          & \multicolumn{1}{p{7.335em}}{KStacks} & 63.64 & 95.15 & \textcolor[rgb]{ .22,  .463,  .114}{1.92} & \textcolor[rgb]{ .8,  0,  0}{-81.82} \\
          & \multicolumn{1}{p{7.335em}}{PrimeStack} & 56.92 & 76.44 & \textcolor[rgb]{ .22,  .463,  .114}{2.05} & \textcolor[rgb]{ .8,  0,  0}{-22.59} \\
          & \multicolumn{1}{p{7.335em}}{KStacksColor} & 80.00 & 11.86 & \textcolor[rgb]{ .8,  0,  0}{-23.14} & \textcolor[rgb]{ .22,  .463,  .114}{13.68} \\
    \midrule
    \multicolumn{1}{p{5.915em}}{Alfred} & \multicolumn{1}{p{7.335em}}{ExplicitInstruct} & 100.00 & 100.00 & {-} & {-} \\
          & \multicolumn{1}{p{7.335em}}{MoveSynonym} & 90.00 & 96.00 & {-} & {-} \\
          & \multicolumn{1}{p{7.335em}}{MoveNextTo} & 100.00 & 100.00 & \textcolor[rgb]{ 1,  .6,  0}{0.00} & \textcolor[rgb]{ .8,  0,  0}{-100.00} \\
          & \multicolumn{1}{p{7.335em}}{MoveToCount2} & 95.83 & 89.58 & \textcolor[rgb]{ .22,  .463,  .114}{4.17} & \textcolor[rgb]{ .8,  0,  0}{-89.58} \\
          & \multicolumn{1}{p{7.335em}}{MoveToCount3} & 90.54 & 83.78 & \textcolor[rgb]{ .8,  0,  0}{-36.69} & \textcolor[rgb]{ .8,  0,  0}{-72.24} \\
          & \multicolumn{1}{p{7.335em}}{MoveToMore} & 97.67 & 93.02 & \textcolor[rgb]{ .8,  0,  0}{-4.81} & \textcolor[rgb]{ .8,  0,  0}{-64.45} \\
          & \multicolumn{1}{p{7.335em}}{MoveNested} & 98.31 & 89.83 & \textcolor[rgb]{ .8,  0,  0}{-8.07} & \textcolor[rgb]{ .8,  0,  0}{-72.76} \\
          & \multicolumn{1}{p{7.335em}}{MoveNested2} & 84.91 & 86.79 & \textcolor[rgb]{ .8,  0,  0}{-38.10} & \textcolor[rgb]{ .8,  0,  0}{-67.64} \\
          & \multicolumn{1}{p{7.335em}}{CutFruits} & 85.00 & 85.00 & {-} & {-} \\
          & \multicolumn{1}{p{7.335em}}{PrepareMeal} & 100.00 & 83.33 & \textcolor[rgb]{ 1,  .6,  0}{0.00} & \textcolor[rgb]{ .8,  0,  0}{-33.33} \\
          & \multicolumn{1}{p{7.335em}}{IceCream} & 100.00 & 93.75 & \textcolor[rgb]{ 1,  .6,  0}{0.00} & \textcolor[rgb]{ .8,  0,  0}{-18.75} \\
          & \multicolumn{1}{p{7.335em}}{SetTable2} & 100.00 & 100.00 & {-} & {-} \\
          & \multicolumn{1}{p{7.335em}}{CleanKitchen} & 94.44 & 94.44 & \textcolor[rgb]{ .8,  0,  0}{-44.44} & \textcolor[rgb]{ .22,  .463,  .114}{5.56} \\ \hline\hline 
    \end{tabular}%
  \label{tbl:erroranalysis}%
\end{table*}%

%% file: conclusion.tex
\section{Conclusion}
\mypara{Summary.} In this paper, we presented an empirical study into the effectiveness of LLMs, specifically GPT-3.5 variants, for the task of natural language goal translation to PDDL. Our key finding was that LLMs have potent linguistic processing ability --- formal linguistic competence~\cite{mahowald2023} --- and can leverage commonsense knowledge (e.g., associations between objects) to achieve impressive translation performance, especially on the common household tasks in ALFRED-L. However, we also find that the models lack fundamental formal and physical world reasoning capabilities that are necessary for translation. This can be problematic for robotics tasks which involve tasks that are less common (e.g., in specialized factories or settings involving assistive care) and not well-represented in the training corpus. Indeed, our work suggests that there is room to improve the LLMs, potentially to integrate reasoning even at the translation stage.

\mypara{Limitations and Future Work.} Our study can be improved in several ways. First, previous work has noted that LLMs can be sensitive to the prompts used. We also found this to be true for goal translation. An exhaustive experiment involving all possible prompts was intractable, so we relied on previous findings~\cite{collins2022,silver2022} and initial trials to design reasonable prompts. However, it is possible that other prompt structures may give better performance. Next, our subtask analysis was necessarily limited given only API access to the models. The subtask tests cannot unambiguously confirm if an LLM succeeds on a given subtask as performance is always contingent upon the LLM being able to output a text answer. We are unable to always pinpoint errors in the transcription of latent information to text and future work can examine latent encodings to derive stronger conclusions. Nevertheless, we hope our results and analysis shed light on the potential of applying LLMs to goal translation.

%% file: ack.tex
\section*{Acknowledgements}
This research is supported by the National Research Foundation, Singapore under its Medium Sized Center for Advanced Robotics Technology Innovation.

%% file: appendix.tex
\clearpage
\appendix
\nobalance

In this section, we provide supplementary material regarding our benchmark problems, evaluation, and prompts for error analysis. We also provide additional results (breakdown according to number of blocks for the Blocksworld domain) and zero-shot results on  the ALFRED-L tasks. Code is available at \texttt{\url{https://github.com/clear-nus/gpt-pddl/}}.

\subsection{Blocksworld benchmark generation} This subsection details how test cases are generated for all tasks on the Blocksworld domain.

\mypara{Problem Generation} Using the PDDL problem generator from \cite{slaney2001}, we first generate PDDL problems with different numbers of objects. There are three options of the number of objects: 4, 8, and 12. Each option has 100 corresponding problems generated. Since these problems use alphabet letters as object symbols, they serve as templates for generating concrete test problems suited for specific tasks. Additional problems are generated as templates for $n$-shot examples.

For each task, the objects in the problem templates are first randomly replaced by the actual blocks. 
Then, a text instruction is sampled for each problem. The original goal generated along with a problem template is removed and does not serve as the ground truth since it cannot satisfy our task requirement most of the time, but it does contribute to the instruction sampling in tasks like ExplicitStacks so as to simplify the pipeline.

\mypara{Instruction Generation} The instructions are constructed with natural language templates. We briefly describe how instructions are sampled in each task below. 
\begin{enumerate}[label=(\alph*)]
    \item \textbf{ExplicitStacks}    
    We first construct a complete block configuration that satisfies the original goal in problem template, then describe the configuration stack by stack. For each stack, we first describe it precisely with predicates, then use language templates to translate this formula form into text. The colors don't repeat.

    \item \textbf{ExplicitStacks-II}
    The sampling process is similar to ExplicitStacks, except that each stack is specified with a sequence of blocks and an indication of the direction, that is, whether the blocks are specified from bottom to top or from top to bottom. Note that we generate one benchmark for each direction, therefore this task has twice as many test cases as most of the other tasks. The colors don't repeat.

    \item \textbf{BlockAmbiguity}
     The sampling process is similar to ExplicitStacks, but for each stack the \texttt{ontable} and \texttt{clear} predicates are not translated, and colors are repeatable.
     
    \item \textbf{NBlocks}
     We first construct a complete block configuration that satisfies the original goal in problem template, then uniformly choose one stack of blocks out of this configuration, and use the number of blocks in this stack as the parameter N in the instruction. The colors don't repeat.

    \item \textbf{KStacks}
    The parameter K is sampled uniformly from non-trivial divisors of the total number of objects defined in problem. The colors don't repeat.

    \item \textbf{PrimeStack}
    The instruction always asks to build a stack of blocks with a prime number of blocks. The colors don't repeat.

    \item \textbf{KStacksColor}
    The parameter K is sampled uniformly from a predefined range, ensuring that the solution exists. The colors can repeat.
\end{enumerate}

We also introduce variations on the verb to make the instructions more diverse. For all phrases like ``create a stack", the verb ``create" is replaced by a synonym like ``make".

It's worth mentioning that since the domain of Blocksworld is short enough, it is treated as part of the query in every example rather than a prefix of the whole prompt. This is not the case in ALFRED-L, since repeating its domain definition would easily exceed the token limit. All tasks generate benchmarks from problems of either all 3 object number options (4, 8, 12) or 2 of them (8, 12). There are 100 test cases per object number option per task. See Tbl. \ref{tbl:bwresults} for the detailed configuration.

\subsection{Blocksworld Evaluation}

All Blocksworld tasks have a common evaluation pipeline that involves the following four stages: 
\begin{enumerate}
    \item \textbf{Problem parsing} fills the generated goal into the prompted incomplete PDDL problem, and use a PDDL dedicated parser to parse the problem text along with the Blocksworld domain file. Then, the goal formula is extracted from the output of parser. This stage performs the domain/syntax error check. If the goal is syntactically incorrect, or an unseen predicate or object is used in the goal, or an predicate has erroneous arguments, the parser will report a \textbf{domain/syntax error}.

    \item \textbf{Graph construction} converts the goal formula into a directed graph. With all predicates in the goal specifying spatial relations between blocks, this stage attempts to build a directed graph that represents whether a block is on another block by existence of a directed edge between the two corresponding nodes. \texttt{table} and \texttt{air} are introduced as two special nodes for encoding \texttt{ontable} and \texttt{clear} predicates. The graph represents at least one physically valid configuration of blocks, which means that no two edges are starting from or pointing to the same node (except that the starting node is \texttt{table} or the ending node is \texttt{air}), and there shall exist neither cycles nor loops. This stage performs the physical error check, and any error in the graph construction process is reported as a \textbf{physical error}.

    \item \textbf{Stack construction} converts the graph into a set of chains of blocks that are accepted as stacks. This stage is also where the loose metric and the strict metric differ. If a chain of blocks in the graph satisfies the criteria of the metric, it is accepted as a valid stack and considered in the last stage. Unaccepted blocks are simply ignored. This stage does not need error checking. The difference between the two metrics is:
    \begin{itemize}
    \item \textbf{Loose metric}: accepts a chain of blocks as a stack as long as its bottom block has the \texttt{ontable} property OR its top block has the \texttt{clear} property. Thus if a block is not mentioned, it is ignored in tje evaluation.
    \item \textbf{Strict metric}: accepts a chain of blocks as a stack as long as its bottom block has the \texttt{ontable} property AND its top block has the \texttt{clear} property.
    \end{itemize}

    \item \textbf{Constraint Validation} evaluates the accepted stacks against the natural language instruction. Evaluation at this stage varies depending on the tasks. If the stacks do not satisfy the instruction, it is reported as a \textbf{constraint violation}. 
\end{enumerate}

\input{tbl_bw_results.tex}

\subsection{ALFRED-L Evaluation}
In the following, we provide a specification of the losse and strict criteria applied for each of the ALFRED-L tasks.

\begin{enumerate}[label=(\alph*)]
\item \textbf{ExplicitInstruct}
\begin{itemize}
    \item Loose: Predicted goal specification includes all predicates in the ground truth goal specification.
    \item Strict: Predicted goal specification matches the ground truth goal specification exactly.
\end{itemize}
\item \textbf{MoveSynonym}
\begin{itemize}
    \item Loose: If the synonym is a receptacle: whether the receptacle receives the moved object. If the synonym is a movable object: whether the object is moved to the correct position. 
    \item Strict: Predicted goal specification has no agent inside. 
\end{itemize}
\item \textbf{MoveNextTo}
\begin{itemize}
    \item Loose: Predicted goal specification includes the two objects at the same location. 
    \item Strict: Predicted goal specification includes the two objects at the same location. The target object is not moved. There is no agent in the goal state. 
\end{itemize}
\item \textbf{MoveToCount2}
\begin{itemize}
    \item Loose:  Predicted goal specification includes the object in the correct box. 
    \item Strict: Predicted goal specification includes the object in the correct box. There is no agent in the goal state. 
\end{itemize}
\item \textbf{MoveToCount3}
\begin{itemize}
    \item Loose:  Predicted goal specification includes the object in the correct box. 
    \item Strict:  Predicted goal specification includes the object in the correct box. There is no agent in the goal state. 
\end{itemize}
\item \textbf{MoveToMore}
\begin{itemize}
    \item Loose: Predicted goal specification includes the object in the correct box.
    \item Strict: Predicted goal specification includes the object in the correct box. There is no agent in the goal state.
\end{itemize}
\item \textbf{MoveNested}
\begin{itemize}
    \item Loose: Predicted goal specification includes the object in the correct box.
    \item Strict: Predicted goal specification includes the object in the correct box. There is no agent in the goal state. There are no redundant predicates in the goal state. 
\end{itemize}
\item \textbf{MoveNested2}
\begin{itemize}
    \item Loose: Predicted goal specification includes the object in the correct box.
    \item Strict: Predicted goal specification includes the object in the correct box. There is no agent in the goal state. There is no redundant predicates in the goal state.
\end{itemize}
\item \textbf{CutFruits} 
\begin{itemize}
    \item Loose: Some fruits are cut and put on the plate.
    \item Strict: Only fruits are cut and put on the plate.
\end{itemize}
\item \textbf{PrepareMeal} 
\begin{itemize}
    \item Loose: Some food is served on the table.
    \item Strict: Only food is served on the table.
\end{itemize}
\item \textbf{IceCream}
\begin{itemize}
    \item Loose: Ice cream is put in the fridge.
    \item Strict: Only ice cream is put in the fridge, with no change to other predicates.
\end{itemize}
\item \textbf{SetTable2} 
\begin{itemize}
    \item Loose: The table is set for two persons.
    \item Strict: The table is set for two persons with no change to other predicates.
\end{itemize}
\item \textbf{CleanKitchen} 
\begin{itemize}
    \item Loose: Nothing on the floor and no violation of the can-contain rule.
    \item Strict: Nothing in the sink. Nothing in the microwave or oven.
\end{itemize}
\end{enumerate}

\subsection{BlocksWorld Error Analysis}

\mypara{Domain Understanding.} For the domain understanding analysis, the five queries are applied in the context of all test cases. The prompts are exactly the same as those used for goal translation, except that the instruction describing the goal state is replaced with one of the five queries (note that instructions and answers in the examples are kept unchanged). We briefly describe the query-answer pair construction processes below:

\begin{itemize}
    \item \textbf{Object Extraction}: there is no randomness in the query. The answer should be all objects listed in the problem definition. The model is asked to split the output objects with spaces.
    \item \textbf{Color-based Object Extraction}: one color in query is uniformly sampled from all colors of blocks that appear in the problem. The answer should be those blocks with this certain color. The model is asked to split the output objects with spaces.
    \item \textbf{On Predicate}: one fact using the \texttt{on} predicate is uniformly sampled from all facts listed in the initial state. Then, with 50\% chance, one of its arguments is replaced by another random block. The query asks about the correctness of the fact. The ground truth answer depends on whether the argument replacement is executed. 
    \item \textbf{OnTable Predicate}: first, we extract all blocks on the table (modified by \texttt{ontable} predicate) in the initial state. Then, the object in query is sampled from these blocks. With 50\% chance, we sample the object in query from blocks that are not on the table instead.
    \item \textbf{Clear Predicate}: Same as the process for \texttt{ontable} predicate, except that we look for the \texttt{clear} predicates in the initial state.
\end{itemize}

Since running these five queries over all test cases would be very costly, we sample 7 out of every 100 test cases for this analysis, except for benchmarks where successes/failures are rare.

\mypara{Goal Inference.} For the goal inference analysis, we use the same domain definition, example problems and test problems as those in the goal translation task, except that all goal specifications (both in the examples and in the test queries) are represented by hierarchical Python lists instead of PDDL formulas. We use list of lists of PDDL object symbols to specify all the stacks in a satisfying configuration. The instructions are also rephrased accordingly. Despite the ambiguity in direction of block sequences under this particular form, we expect the model to discriminate the bias underneath when required (note that many tasks actually accept interpretations in both directions), provided that answers in the examples are also replaced by corresponding lists. 

\subsection{ALFRED-L Error Analysis}

\input{error_analysis_zero_shot_tbl.tex}

\begin{table}
\centering
\caption{Success Rates on the Goal Translation Tasks (Zero Shot)}
\label{tbl:results_zero_shot_success}
\begin{tabular}{llrr}
\hline\hline
                &                    & \multicolumn{2}{c}{\textbf{Success Rate (\%)}}                           \\
\textbf{Domain} & \textbf{Task Name} & \multicolumn{1}{r}{\textbf{Loose}} & \multicolumn{1}{r}{\textbf{Strict}} \\ \hline

ALFRED-L                     & MoveSynonym        & 98.00                              & 90.00                               \\
                & MoveNextTo         & 96.00                              & 95.00                               \\
                & MoveToCount2       & 78.00                              & 72.00                               \\
                & MoveToCount3       & 47.00                              &43.00                               \\
                & MoveToMore         & 73.00                              & 69.00                               \\
                & MoveNested         & 58.00                              & 55.00                               \\
                & MoveNested2        & 43.00                              & 39.00                               
                           \\ \hline\hline
\end{tabular}
\end{table}

ALFRED-L errors were analyzed via direct questions regarding the domain and/or goal. 

\begin{enumerate}[label=(\alph*)]
\item \textbf{ExplicitInstruct}
\begin{itemize}
    \item Domain understanding: Where is the \{object\}?
    \item Goal inference: Write the goal specification in natural language. 
\end{itemize}
\item \textbf{MoveSynonym}. There are two cases, as the synonym can be either a receptacle or an object. If it is an object: 
\begin{itemize}
    \item Domain understanding: Is there a \{synonym\} in the initial state?
    \item Goal inference: Put a \{synonym\} on Desk1. Which object should we put on Desk1?
\end{itemize}
If it is a receptacle:
\begin{itemize}
    \item Domain understanding: Is there a \{synonym\} in the initial state?
    \item Goal inference: Put the \{random\_object\} on/in a \{synonym\}. Where should we put \{random\_object\}?
\end{itemize}
\item \textbf{MoveNextTo}
\begin{itemize}
    \item Domain understanding: Where is \{target\_object\} in the initial state?
    \item Goal inference: Put \{moved\_object\} next to \{target\_object\}. Do not move \{target\_object\}. Where should we put \{moved\_object\}?
\end{itemize}
\item \textbf{MoveToCount2}
\begin{itemize}
    \item Domain understanding: Which box has two \{target\_type\}s in the initial state?
    \item Goal inference: Move \{moved\_object\} to the box with two \{target\_type\}s. Which object should we move {\{moved\_object\}} into?
\end{itemize}
\item \textbf{MoveToCount3}
\begin{itemize}
    \item Domain understanding: Which box has three \{target\_type\}s in the initial state?
    \item Goal inference: Move \{moved\_object\} to the box with three \{target\_type\}s. Which object should we move {\{moved\_object\}} into?
\end{itemize}
\item \textbf{MoveToMore}
\begin{itemize}
    \item Domain understanding: Which box has more \{target\_type\}s in the initial state?
    \item Goal inference: Move \{moved\_object\} to the box with more \{target\_type\}s. Which object should we move {\{moved\_object\}} into?
\end{itemize}
\item \textbf{MoveNested}
\begin{itemize}
    \item Domain understanding: Which sofa is \{target\_object\} in?
    \item Goal inference: Put the \{moved\_object\} on the sofa with \{target\_object\}. Do not put it in box. Which object should we put \{moved\_object\} on?
\end{itemize}
\item \textbf{MoveNested2}
\begin{itemize}
    \item Domain understanding: Which sofa is \{target\_object\} in?
    \item Goal inference: Put the \{moved\_object\} on the sofa with \{target\_object\}. Do not put it in box. Which object should we put \{moved\_object\} on?
\end{itemize}
\item \textbf{CutFruits}
\begin{itemize}
    \item Domain understanding: What are the fruits?
    \item Goal inference: Write the goal specification in natural language. 
\end{itemize}
\item \textbf{PrepareMeal}
\begin{itemize}
    \item Domain understanding: What does ``prepare a meal'' mean? What are the relevant objects for preparing a meal?
    \item Goal inference: Write the goal specification in natural language. 
\end{itemize}
\item \textbf{IceCream}
\begin{itemize}
    \item Domain understanding: Where does the ice cream belong?
    \item Goal inference: Write the goal specification in natural language. 
\end{itemize}
\item \textbf{SetTable2}
\begin{itemize}
    \item Domain understanding: What does ``set the table'' mean? What are the relevant objects for setting the table?
    \item Goal inference: Write the goal specification in natural language. 
\end{itemize}
\item \textbf{CleanKitchen}
\begin{itemize}
    \item Domain understanding: What does ``clean up the kitchen'' mean? What are the objects relevant objects for cleaning up the kitchen? Where to put them?
    \item Goal inference: Write the goal specification in natural language. 
\end{itemize}
\end{enumerate}

\subsection{
Zero-shot performance for ALFRED-L partially specified tasks}
The zero-shot performance for ALFRED-L partially specified tasks is listed in Table \ref{tbl:results_zero_shot_success}. An error analysis is shown in Table \ref{tbl:error_zero_shot}. In general, performance was lower in the zero-shot setting than in the one-shot setting.

\subsection{
Fine-grained Performance Report on Blocksworld Tasks}
Tbl. \ref{tbl:bwresults} lists detailed success rates and error rates of translation results on all Blocksworld tasks. Results of each task are split by the number of objects appearing in the problem and evaluated separately. For the ExplicitStacks-II task, the benchmark is split into two parts based on the direction of the block sequence in the instruction (whether in each stack the firt block mentioned is the one at bottom or the one on the top). The constraint violation rates are based on the strict metric. 
In general, the model makes few PDDL domain-related errors (i.e., PDDL syntax check related to use of correct syntax and existence of predicates and objects), which reflects that it understands the syntax of PDDL and knows the objects it can use for building a stack. In contrast, the model fails to avoid physical errors, especially in tasks that involve specification of multiple stacks. This coincides with the results from the domain understanding subtask tests (in Sec. \ref{sec:ResultsAndDiscussion}) where the model achieved high performance in object extraction tests but performs much worse in predicate understanding tests.

\input{tbl_bw_pred_order}

\subsection{ExplicitStacks-II: Sensitivity to Sequential Patterns}

In the ExplicitStacks-II task on Blocksworld, there is a significant performance drop when switching the order of block specification in stacks from the bottom-to-top pattern to the top-to-bottom pattern. One hypothesis for this phenomenon is that the PDDL goal in the example always specifies first the block at the bottom of a stack with an \texttt{ontable} predicate, then a sequence of \texttt{on} predicates, and last the block at the top with a \texttt{clear} predicate, which matches a bottom-to-top pattern. Hence, the model captures the linguistic relation between the PDDL goal and text instruction more easily when they have the same pattern. To validate this hypothesis, we provide a different example for the benchmark with top-to-bottom pattern. This example uses exactly the same PDDL problem as the original one, but reverses the order of predicates in specification of every stack when defining the ground truth PDDL goal. This example matches a top-to-bottom pattern. As a result, the performance is improved by a significant margin as shown in Tbl. \ref{tbl:bwpred}. This implies that the model is sensitive to the sequential patterns in $n$-shot examples.

\input{tbl_bw_nshot}

\subsection{KStacks: Do More Examples Guarantee Generalization?}

In the KStacks task on Blocksworld, one can see that the performance decreases as the number of objects increases. Considering that the number of stacks requested in the instruction is uniformly sampled from the non-trivial divisors of total number of objects, an increase in objects leads to more diverse instructions, which makes it more difficult for the model to choose the right length of block sequences. In fact, a closer observation on the generated goals shows that the model is copying the goal pattern (including the number of stacks and the height of each stack) in 90-95\% cases where number of objects is 8 and in 50-60\% cases where number of objects is 12. Potentially, the model may perform better if more examples are provided to help it generalize. However, this turns out not be the case. Even provided with a second example where the instruction requests a different number of stacks, there is no significant improvement, as shown in Tbl. \ref{tbl:bwnshot}. The table lists all the values of $K$ (the number of stacks to build) used in both $n$-shot ($n=1, 2$) examples and test cases. Our results show that even when all possible answer patterns are provided (as in the 8-object cases), the model still failed to reliably translate instructions. Possibly, the $n$-shot prompt design may need to be improved for translation.  

%% file: tbl_bw_results.tex
\begin{table*}[htbp]
\centering
\caption{Success Rates and Error Report on the Blocksworld Goal Translation Tasks}
\label{tbl:bwresults}
\begin{tabular}{lrrrrrr}
\toprule
\toprule
\textbf{}                         & \multicolumn{1}{l}{\textbf{}} & \multicolumn{2}{c}{\textbf{Success Rates (\%)}}                          & \multicolumn{3}{c}{\textbf{Detailed Error Rates (\%)}}                                                                                             \\
\textbf{Task Name}                & \textbf{Number of Objects}    & \multicolumn{1}{r}{\textbf{Loose}} & \multicolumn{1}{r}{\textbf{Strict}} & \multicolumn{1}{c}{\textbf{Domain/Syntax Error}} & \multicolumn{1}{c}{\textbf{Physical Error}} & \multicolumn{1}{c}{\textbf{Constraint Violation}} \\ \midrule\midrule
ExplicitStacks                    & 4                             & 100.00                             & 100.00                              & 0.00                                             & 0.00                                        & 0.00                                              \\
                                  & 8                             & 100.00                             & 100.00                              & 0.00                                             & 0.00                                        & 0.00                                              \\
                                  & 12                            & 99.00                              & 96.00                               & 0.00                                             & 1.00                                        & 3.00                                              \\
Average                           & \multicolumn{1}{l}{}          & 99.67                              & 98.67                               & \multicolumn{1}{l}{}                             & \multicolumn{1}{l}{}                        & \multicolumn{1}{l}{}                              \\ \midrule
ExplicitStacks-II (bottom to top) & 4                             & 55.00                              & 48.00                               & 0.00                                             & 13.00                                       & 39.00                                             \\
                                  & 8                             & 98.00                              & 77.00                               & 0.00                                             & 1.00                                        & 22.00                                             \\
                                  & 12                            & 99.00                              & 99.00                               & 0.00                                             & 1.00                                        & 0.00                                              \\
Average                           & \multicolumn{1}{l}{}          & 84.00                              & 74.67                               & \multicolumn{1}{l}{}                             & \multicolumn{1}{l}{}                        & \multicolumn{1}{l}{}                              \\ \midrule
ExplicitStacks-II (top to bottom) & 4                             & 19.00                              & 17.00                               & 0.00                                             & 17.00                                       & 66.00                                             \\
                                  & 8                             & 35.00                              & 30.00                               & 0.00                                             & 3.00                                        & 67.00                                             \\
                                  & 12                            & 42.00                              & 42.00                               & 0.00                                             & 0.00                                        & 58.00                                             \\
Average                           & \multicolumn{1}{l}{}          & 32.00                              & 29.67                               & \multicolumn{1}{l}{}                             & \multicolumn{1}{l}{}                        & \multicolumn{1}{l}{}                              \\ \midrule
NBlocks                           & 4                             & 72.00                              & 62.00                               & 0.00                                             & 2.00                                        & 36.00                                             \\
                                  & 8                             & 50.00                              & 46.00                               & 0.00                                             & 5.00                                        & 49.00                                             \\
                                  & 12                            & 64.00                              & 64.00                               & 0.00                                             & 0.00                                        & 36.00                                             \\
Average                           & \multicolumn{1}{l}{}          & 62.00                              & 57.33                               & \multicolumn{1}{l}{}                             & \multicolumn{1}{l}{}                        & \multicolumn{1}{l}{}                              \\ \midrule
KStacks                           & 4                             & 94.00                              & 94.00                               & 0.00                                             & 5.00                                        & 1.00                                              \\
                                  & 8                             & 51.00                              & 51.00                               & 1.00                                             & 2.00                                        & 46.00                                             \\
                                  & 12                            & 20.00                              & 20.00                               & 0.00                                             & 12.00                                       & 68.00                                             \\
Average                           & \multicolumn{1}{l}{}          & 55.00                              & 55.00                               & \multicolumn{1}{l}{}                             & \multicolumn{1}{l}{}                        & \multicolumn{1}{l}{}                              \\ \midrule
PrimeStack                        & 8                             & 97.00                              & 97.00                               & 0.00                                             & 0.00                                        & 3.00                                              \\
                                  & 12                            & 77.00                              & 77.00                               & 0.00                                             & 0.00                                        & 23.00                                             \\
Average                           & \multicolumn{1}{l}{}          & 87.00                              & 87.00                               & \multicolumn{1}{l}{}                             & \multicolumn{1}{l}{}                        & \multicolumn{1}{l}{}                              \\ \midrule
BlockAmbiguity                    & 8                             & 26.00                              & 23.00                               & 2.00                                             & 17.00                                       & 58.00                                             \\
                                  & 12                            & 10.00                              & 6.00                                & 0.00                                             & 25.00                                       & 69.00                                             \\
Average                           & \multicolumn{1}{l}{}          & 18.00                              & 14.50                               & \multicolumn{1}{l}{}                             & \multicolumn{1}{l}{}                        & \multicolumn{1}{l}{}                              \\ \midrule
KStacksColor                      & 8                             & 7.00                               & 7.00                                & 1.00                                             & 16.00                                       & 76.00                                             \\
                                  & 12                            & 31.00                              & 31.00                               & 0.00                                             & 4.00                                        & 65.00                                             \\ 
Average                           & \multicolumn{1}{l}{}          & 19.00                              & 19.00                               & \multicolumn{1}{l}{}                             & \multicolumn{1}{l}{}                        & \multicolumn{1}{l}{}   \\ \hline \hline                          
\end{tabular}
\end{table*}

%% file: error_analysis_zero_shot_tbl.tex
\begin{table*}[hbt!]
\centering
\caption{Error Analysis on Translation Subtasks across Tasks (Zero Shot).}
\label{tbl:error_zero_shot}
\begin{tabular}{llllrrr}
\hline\hline
                &                    &                        & \multicolumn{2}{c}{\textbf{Error Rate (\%) under Successes}}                                     & \multicolumn{2}{c}{\textbf{Error Rate (\%) under Failure}}                                       \\
\textbf{Domain} & \textbf{Task Name} & \textbf{Num. Failures} & \multicolumn{1}{c}{\textbf{Domain Understanding}} & \multicolumn{1}{c}{\textbf{Goal Inference}} & \multicolumn{1}{c}{\textbf{Domain Understanding}} & \multicolumn{1}{c}{\textbf{Goal Inference}} \\ \hline

ALFRED-L          
           & MoveSynonym        & 2/100                 &    \multicolumn{1}{r}{48.98}                                               &  \multicolumn{1}{r}{1.02}                         & 100.00                                             & 0.00                                         \\
                & MoveNextTo         & 4/100                  &     \multicolumn{1}{r}{0.00}                                              &  \multicolumn{1}{r}{5.21}                         & 0.00                                              & 75.00                                        \\
                & MoveToCount2       & 22/100                 &    \multicolumn{1}{r}{0.00}                                               &  \multicolumn{1}{r}{2.56}                         & 0.00                                              & 52.17                                        \\
                & MoveToCount3       & 53/100                 & \multicolumn{1}{r}{17.02}                                                  &  \multicolumn{1}{r}{8.51}                         & 16.98                                             & 86.79                                        \\
                & MoveToMore         & 27/100                 &    \multicolumn{1}{r}{1.37}                                               &  \multicolumn{1}{r}{2.74}                         & 11.11                                             & 59.26                                        \\
                & MoveNested         & 42/100                 &   \multicolumn{1}{r}{5.17}                                                &   \multicolumn{1}{r}{8.62}                         & 4.76                                              & 80.95                                        \\
                & MoveNested2        & 57/100                 &   \multicolumn{1}{r}{11.63}                                                & \multicolumn{1}{r}{11.63}                         & 31.58                                             & 85.96                                                          \\\hline\hline
\end{tabular}
\end{table*}

%% file: tbl_bw_pred_order.tex
\begin{table}
\centering
\caption{ExplicitStacks-II (top to bottom): \\Effect of the Example Predicate Order}
\label{tbl:bwpred}
\begin{tabular}{lrrr}
\toprule
\toprule
                         & \multicolumn{1}{l}{}       & \multicolumn{2}{c}{\textbf{Success Rates (\%)}}                          \\
\textbf{Predicate Order} & \textbf{Number of Objects} & \multicolumn{1}{c}{\textbf{Loose}} & \multicolumn{1}{c}{\textbf{Strict}} \\ \midrule\midrule
OnTable first            & 4                          & 19.00                              & 17.00                               \\
                         & 8                          & 35.00                              & 30.00                               \\
                         & 12                         & 42.00                              & 42.00                               \\
Average                  & \multicolumn{1}{l}{}       & 32.00                              & 29.67                               \\ \midrule
Clear first              & 4                          & 62.00                              & 48.00                               \\
                         & 8                          & 98.00                              & 88.00                               \\
                         & 12                         & 98.00                              & 98.00                               \\
Average                  & \multicolumn{1}{l}{}       & 86.00                              & 78.00                              
                \\ \hline \hline
\end{tabular}
\end{table}

%% file: tbl_bw_nshot.tex
\begin{table}[htbp]
\centering
\caption{KStacks: Effect of the Number of Examples}
\label{tbl:bwnshot}
\begin{tabular}{lrrrrr}
\toprule
\toprule
                    & \multicolumn{1}{l}{}       & \multicolumn{2}{c}{\textbf{\# of Stacks(K)}} & \multicolumn{2}{c}{\textbf{Success Rates(\%)}} \\
\textbf{Prompt} & \textbf{\# of Objects} & \textbf{Example}        & \textbf{Test}          & \textbf{Loose}         & \textbf{Strict}        \\ \midrule\midrule
1-shot              & 4                          & 2                       & 2                      & 94.00                  & 94.00                  \\
                    & 8                          & 4                       & 2, 4                   & 51.00                  & 51.00                  \\
                    & 12                         & 4                       & 2, 3, 4, 6             & 20.00                  & 20.00                  \\
Average             & \multicolumn{1}{l}{}       & \multicolumn{1}{l}{}    & \multicolumn{1}{l}{}   & 55.00                  & 55.00                  \\ \midrule
2-shot              & 4                          & 2, 2                    & 2                      & 100.00                 & 100.00                 \\
                    & 8                          & 2, 4                    & 2, 4                   & 51.00                  & 51.00                  \\
                    & 12                         & 2, 4                    & 2, 3, 4, 6             & 21.00                  & 21.00                  \\
Average             & \multicolumn{1}{l}{}       & \multicolumn{1}{l}{}    & \multicolumn{1}{l}{}   & 57.33                  & 57.33             \\ \hline \hline    
\end{tabular} 
\end{table}